\documentclass[letterpaper,10pt,conference]{ieeeconf}
\IEEEoverridecommandlockouts
\usepackage[T1]{fontenc}
\usepackage{mathptmx}
\usepackage{graphicx,amsmath,amssymb,booktabs,array,multirow}
\usepackage[table]{xcolor}
\usepackage{cite,url,capt-of}
\usepackage{makecell}
\makeatletter
\let\NAT@parse\undefined
\makeatother
\usepackage[hidelinks]{hyperref}

\definecolor{figureink}{HTML}{20304A}
\definecolor{selectedink}{HTML}{008E91}

\definecolor{ForestGreen}{rgb}{0.13, 0.65, 0.13}
\definecolor{fgreen}{rgb}{0.13, 0.75, 0.13}
\definecolor{WildStrawberry}{rgb}{1.0, 0.26, 0.64}
\definecolor{search}{rgb}{0.92, 0.22, 0.05}
\definecolor{static}{rgb}{1.0, 0.67, 0.25}
\definecolor{dynamic}{rgb}{0.93, 1.0, 0.25}
\definecolor{trajectory}{rgb}{0.73, 0.97, 1.0}
\newcommand{\increase}[1]{\fontsize{6pt}{0.5em}\selectfont\color{ForestGreen}{$\uparrow$~\textbf{#1}}}
\newcommand{\decrease}[1]{\fontsize{6pt}{0.5em}\selectfont\color{WildStrawberry}{$\downarrow$~\textbf{#1}}}

\graphicspath{{figures/}}
\title{\LARGE\bf ReInsVLN: Revisiting Historical Evidence for Training-Free Vision-and-Language Navigation}
\author{\fontsize{10}{11}\selectfont
Siwei Cheng$^{1,3,*}$, Hao Wu$^{2,3,*}$, Zitong Qi$^{3}$, Xinming Zhang$^{1,3}$, Haoyu Xi$^{2,3}$,\\
Peiran Yin$^{2,3}$, Haozhe Hu$^{2,3}$, Xudong Wang$^{3}$, Xiaoyu Shen$^{3,\dagger}$, Wei Zhang$^{3,\dagger}$
\thanks{$^{1}$University of Science and Technology of China, $^{2}$Shanghai Jiao Tong University, $^{3}$Eastern Institute of Technology, Ningbo}
\thanks{$^{*}$Equal Contribution, $^{\dagger}$Corresponding Authors: Xiaoyu Shen, Wei Zhang (email: \{xyshen, zhw\}@eitech.edu.cn)}
}

\begin{document}
\maketitle
\thispagestyle{empty}
\pagestyle{empty}

\begin{abstract}
Vision-and-language navigation (VLN) relies on retaining and using relevant observations. Yet growing histories introduce redundancy, and retained information may remain underused. We introduce ReInsBench, a diagnostic benchmark that examines history utilization through controlled continuation trials. Its results show that history-review reinstruction can recover navigation failures using the existing task context. Motivated by this finding, we propose ReInsVLN, a training-free framework supporting memory construction and utilization in frozen navigators. Coverage- and Instruction-Aware Compression constructs compact visual context through temporal redundancy reduction, coverage-aware protection, and instruction-guided token selection. Forgetting-Inspired History Reminders guide navigators to revisit retained observations according to a predefined schedule, without additional model calls. On R2R Val-Unseen, ReInsVLN improves success rates by 8.48 and 3.37 percentage points over the native StreamVLN and InternVLA-N1, reaching 64.27\% and 64.44\%, respectively. Visual compression largely preserves navigation performance while reducing prefill latency and KV cache memory by up to 21.77\% and 32.25\%, respectively, on matched observation workloads. Real-world trials on humanoid and quadruped robots further support the applicability of ReInsVLN to sequential instruction following and ordinal navigation.
\end{abstract}

\section{Introduction}

\begin{figure}[t]
\centering
\includegraphics[width=\columnwidth]{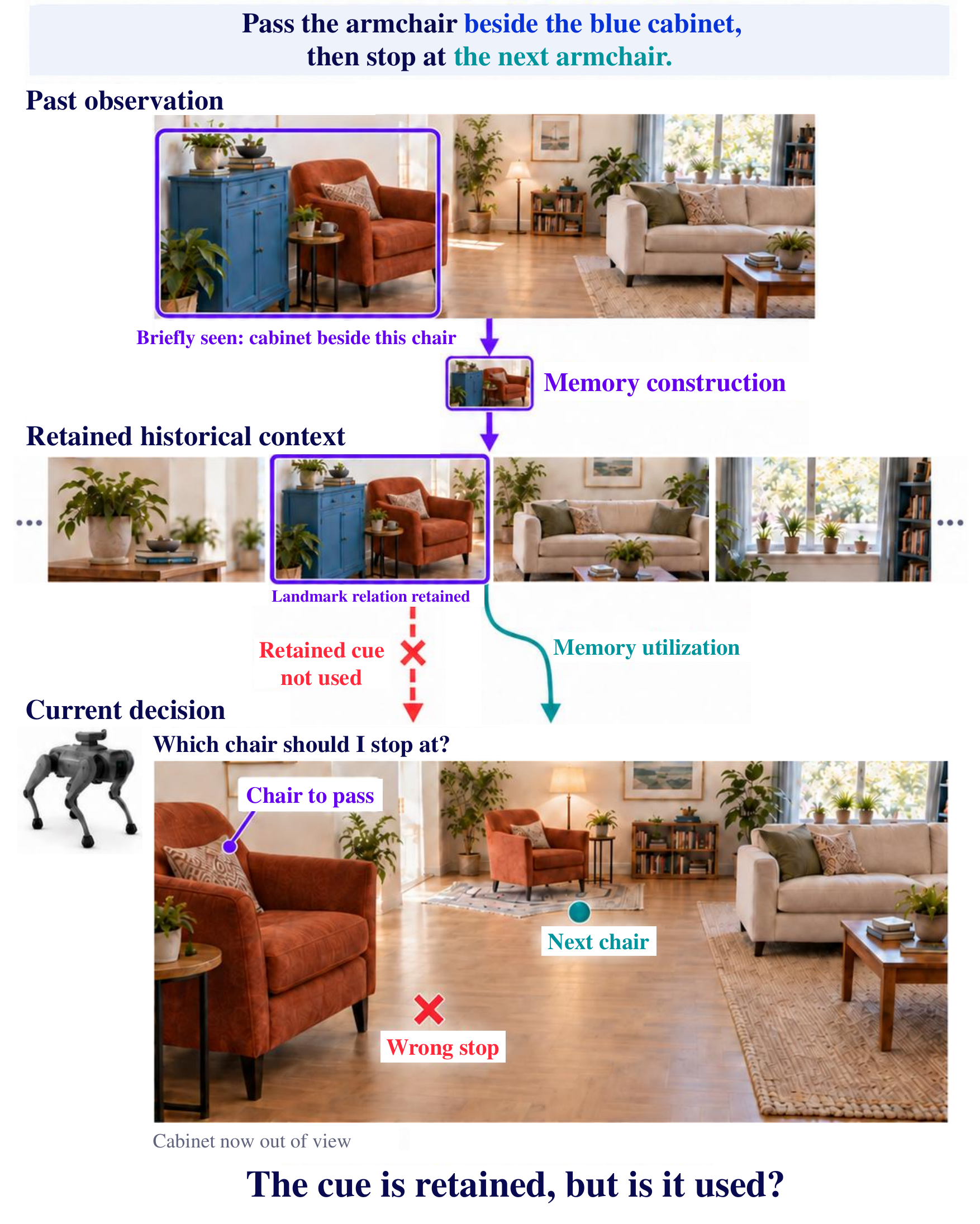}

\caption{Navigation memory requires retaining and using relevant observations. Although the blue cabinet is no longer visible, its retained relation to the first armchair helps identify the next armchair as the stopping point.}
\label{fig:motivation}
\end{figure}

Vision-and-language navigation (VLN) requires an agent to follow natural-language instructions using a sequence of visual observations~\cite{anderson2018r2r, krantz2020vlnce}. Partial observability makes decisions depend on previously encountered landmarks~\cite{hong2021vlnbert}. As illustrated in Fig.~\ref{fig:motivation}, following the instruction ``Pass the armchair beside the blue cabinet, then stop at the next armchair'' requires the navigator to use the previously observed relation between the cabinet and the first armchair to identify the next armchair as the stopping point. Such dependencies make memory essential for retaining information from past observations and using it to guide subsequent actions~\cite{chen2021hamt}.

Effective navigation memory involves two complementary capabilities: \emph{memory construction} and \emph{memory utilization}. Here, memory construction selects and organizes past observations into available context, while memory utilization uses that context to guide decisions. Spatial maps offer one way to organize navigation history~\cite{chen2022duet}. As navigation proceeds, repeated views introduce redundancy into the growing observation history, increasing the cost of processing it~\cite{zhang2024navid, wei2026streamvln, internnav2025, wei2026ground}. This requires retaining useful historical information within limited context and computation budgets, while preserving cues for later decisions. However, retained information does not guarantee effective use, even in long-context language models~\cite{liu2024lostmiddle}. In the example in Fig. 1, the cabinet observation may remain in the navigator's history without guiding it to stop at the correct armchair. This raises a central question: \emph{even when relevant historical information is retained, can the navigator use it effectively?}

To examine this question, we introduce \textbf{ReInsBench}, a diagnostic benchmark constructed from selected R2R~\cite{anderson2018r2r} navigation episodes. At selected decision points, a frozen navigator continues under different prompts from the same observation history and environment state. The benchmark assesses whether directing the navigator to revisit its existing history improves subsequent navigation, using \emph{neutral reminders} to control for the effect of generic prompting. In a controlled subset of 198 instances, history review yields 64 successful continuations, compared with 53 for neutral reminders and 40 without an additional prompt. These findings suggest that \emph{navigators do not always make effective use of the historical information already available to them.}

Building on this diagnosis, we aim to improve memory construction and utilization within existing navigators. A compact visual context must preserve cues whose relevance may emerge only later~\cite{lu2026stepnav, wang2026historypruning}. Once retained, these cues need to inform subsequent decisions at appropriate moments, requiring guidance that encourages their use while limiting interference with otherwise correct navigation.

To this end, we propose \textbf{ReInsVLN}, a training-free framework that combines visual context compression and history-review reminders within the existing context mechanisms of navigators. For memory construction, we adopt a two-stage visual compression strategy consisting of pre-LLM compression and in-LLM compression. The pre-LLM stage reduces temporal redundancy before visual tokens enter the language model, combining causal token reduction with dynamic-token protection to prioritize content poorly covered by previously retained tokens. The in-LLM stage builds a query from the original instruction and expands it with high-scoring visual seeds. It then combines semantic relevance with positional information to select a compact subset of visual tokens, restoring their original order after selection. For memory utilization, short reminders guide the navigator to revisit retained observations in relation to the original instruction. A predefined, forgetting-inspired schedule uses the number of decision calls elapsed since the previous reminder as a temporal prior for determining when to introduce history review. The prompts are inserted at existing decision calls, preserving the native action-execution pipeline.

Experiments on R2R Val-Unseen evaluate ReInsVLN with StreamVLN and InternVLA-N1, representing streaming and non-streaming inference, respectively. With Coverage- and Instruction-Aware Compression and Forgetting-Inspired History Reminders applied together, ReInsVLN improves success rates by 8.48 and 3.37 points over the corresponding native navigators. These improvements exceed the sum of individual gains on both navigators, suggesting synergy between two components. Visual compression also reduces prefill latency and KV cache usage. Physical deployment is another focus of legged-robot VLN~\cite{cheng2025navila}.

The main contributions are summarized as follows:
\begin{itemize}
    \item We introduce ReInsBench to diagnose history utilization through controlled prompting comparisons, showing that history-review reinstruction can recover navigation failures using the existing task context.
    \item We develop Coverage- and Instruction-Aware  Compression for memory construction in ReInsVLN, combining temporal redundancy reduction with instruction-guided visual token selection.
    \item  We introduce Forgetting-Inspired History Reminders to support memory utilization by guiding frozen navigators to revisit retained observations according to a predefined schedule, without additional model calls.
    \item {Simulation experiments show navigation gains on both navigators, together with reduced prefill latency and KV cache storage. Real-world trials on humanoid and quadruped robots further support the training-free generalization of ReInsVLN to challenging tasks requiring sequential landmark tracking.}
\end{itemize}

\section{Related Work}
\subsection{History Modeling in VLN}

History modeling connects current observations with previously encountered landmarks. Explicit visual-language maps~\cite{wen2025ovlmap} organize observations spatially, while NaVid incorporates historical observations into action prediction. StreamVLN separates recent interactions from slowly updated visual memory and reuses cached states across decisions. ReInsBench complements these approaches by assessing history utilization through controlled prompting. ReInsVLN builds on existing context mechanisms, combining visual compression and history-review reminders to reduce redundancy and improve history utilization.

\subsection{Visual Token Compression}

Visual token compression accelerates MLLM inference by shortening visual token sequences, either in a single step or progressively across multiple stages~\cite{wu2026data,yin2026bifocal,zheng2026geogr}. LLaVA-PruMerge~\cite{shang2025llava} prunes and merges tokens before the LLM. FastV~\cite{chen2024fastv} prunes at a selected LLM layer, while PyramidDrop~\cite{xing2024pyramiddrop} and HiDrop~\cite{wu2026hidrop} prune progressively across layers. For streaming video, StreamForest~\cite{zeng2026streamforest} uses hierarchical event-level merging, and ViCoStream~\cite{tan2026vicostream} uses chunk-level temporal token dropping. In VLN, STEP-Nav~\cite{lu2026stepnav} and history-conditioned pruning~\cite{wang2026historypruning} reduce spatial and temporal redundancy. While prior compression work primarily targets inference efficiency, ReInsVLN couples two-stage compression with history-review reminders to improve the utilization of retained visual context.

\subsection{Memory Utilization and Reactivation}

Memory utilization connects retained information with current decisions. CMMR-VLN~\cite{li2026cmmrvln} retrieves prior navigation cases as rules, while Memoir~\cite{xu2026memoir} queries observation and behavioral histories using predicted states. MemVR~\cite{zou2025memvr} reinjects visual evidence into intermediate layers under uncertainty, and VisRef~\cite{ghosal2026visref} reintroduces selected visual tokens during reasoning. MemoryBank~\cite{zhong2024memorybank} and DART-VLN~\cite{zhang2026dartvln} incorporate temporal priors into memory retention and readout, respectively. ReInsVLN provides reactivation at the input level through scheduled history-review prompts, guiding frozen navigators to revisit existing observations without additional retrieval or model calls.

\section{ReInsBench: Diagnosing History Utilization}
\label{sec:diagnostic}

Retaining history does not guarantee its effective use. ReInsBench is a diagnostic benchmark that examines this gap through matched R2R continuation trials. A frozen navigator resumes from the same observation history and environment state under different prompting conditions. Paired outcomes quantify navigation recovery and disruption.

\subsection{Benchmark Construction}

\begin{figure}[t]
\centering
\includegraphics[width=\columnwidth]{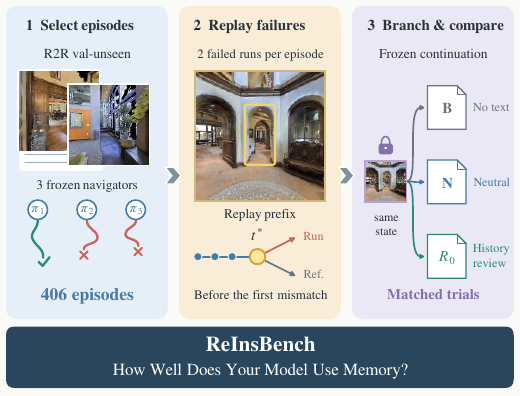}
\caption{Construction of ReInsBench.
Episodes solved by exactly one of three frozen navigators
contribute two unsuccessful rollouts each.
Recorded prefixes are replayed to the state before the first
disagreement with reference actions, yielding 812 model-specific
checkpoints from 406 episodes.
Matched continuations from each checkpoint compare neutral
prompting ($N$) and history review with task restatement ($R_0$);
a 198-instance subset additionally includes controlled
continuation without added text ($B$).}
\label{fig:reinsbench_construction}
\end{figure}

ReInsBench is constructed from R2R Val-Unseen in continuous environments, using recorded rollouts of StreamVLN, InternVLA-N1, and AwareVLN. As illustrated in Fig.~\ref{fig:reinsbench_construction}, episodes are retained only when exactly one navigator succeeds. The successful rollout demonstrates episode-level solvability, while the two unsuccessful rollouts provide diagnostic instances. This yields 812 model-specific instances from 406 episodes, spanning 296 scene--trajectory pairs across 11 scenes.

For each unsuccessful rollout, the first disagreement between recorded and reference actions defines an offline checkpoint. The recorded prefix is replayed to restore the simulator state, with the checkpoint registered before the disagreeing action is executed. The collection contains 451 checkpoints after navigation begins and 361 initial checkpoints, allowing separate analyses of navigation with prior history and initial instruction grounding.

\subsection{Intervention and Evaluation Protocol}

\textbf{Controlled continuation ($B$)} adds no text and controls for renewed inference. \textbf{Neutral prompting ($N$)} requests careful continuation. \textbf{History-review reinstruction ($R_0$)} asks the navigator to revisit available history and realign subgoals with the original instruction. The intervention combines a review request with task restatement. For 794 template reminders, reference sequence length determines a coarse starting clause and the repeated instruction suffix; 18 pilot reminders use fixed semantic restatements. Reference information informs offline construction, but no new reference action sequence is supplied to the navigator.

\begin{table}[t]
\centering

\caption{Matched continuation success counts on ReInsBench. $B$ is evaluated only on the 198-instance subset. The final two rows summarize all 812 instances.}
\label{tab:diagnostic}
\begin{tabular}{lrrrr}
\toprule
\textbf{Diagnostic set} & \textbf{$n$} & \textbf{$B$} & \textbf{$N$} & \textbf{$R_0$} \\
\midrule
Three-branch subset & 198 & 40 & 53 & 64 \\
Subset with prior navigation & 148 & 37 & 47 & 51 \\
Complete two-branch set & 812 & -- & 190 & 209 \\
\midrule
\multicolumn{2}{l}{$N$ only / $R_0$ only}
  & \multicolumn{3}{r}{82 / 101} \\
\multicolumn{2}{l}{Both succeed / either succeeds}
  & \multicolumn{3}{r}{108 / 291} \\
\bottomrule
\end{tabular}
\end{table}

At each checkpoint, the frozen navigator resumes from the same replayed state under different prompting conditions. We evaluate whether each continuation reaches the goal and compare the resulting successes and failures. Comparisons with $B$ assess the benefit of reinstruction over unprompted continuation, while comparisons with $N$ assess its additional benefit over a generic reminder. The full benchmark contains 812 instances evaluated under $N$ and $R_0$, including a subset of 198 instances evaluated under all three conditions.

\begin{figure}[!b]
\centering
\includegraphics[width=\linewidth]{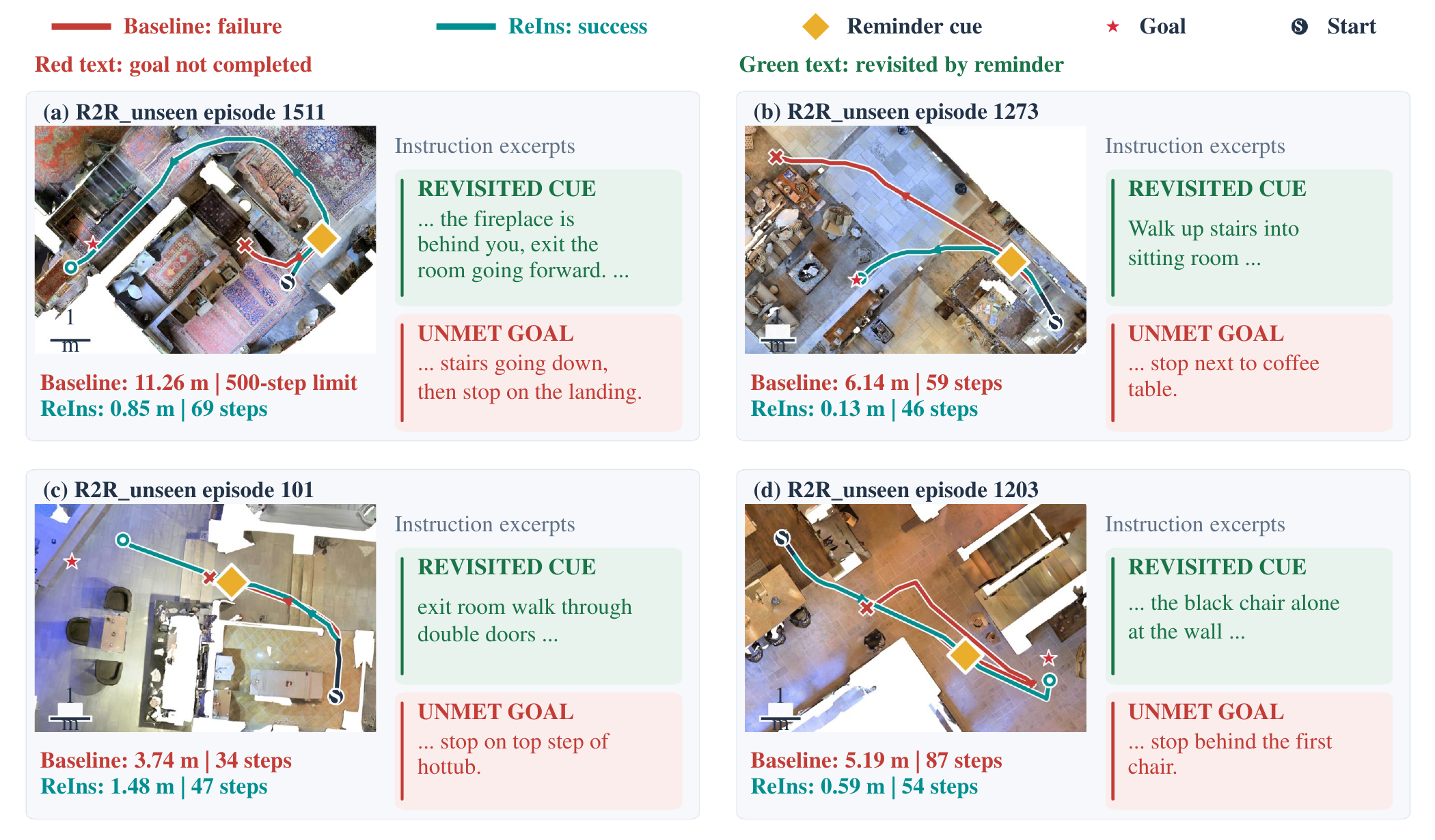}
\caption{Selected StreamVLN continuations on ReInsBench: $B/N$ fail and $R_0$ succeeds. Habitat-Sim replays show final navigation errors alongside original instruction excerpts. Green and red spans mark revisited cues and unmet goals, ellipses mark omissions. Maps, cue diamonds, and goal stars are for illustration only, not navigator inputs or causal labels.}
\label{fig:bench_examples}
\end{figure}

\begin{figure*}[!t]
\centering
\includegraphics[width=\textwidth]{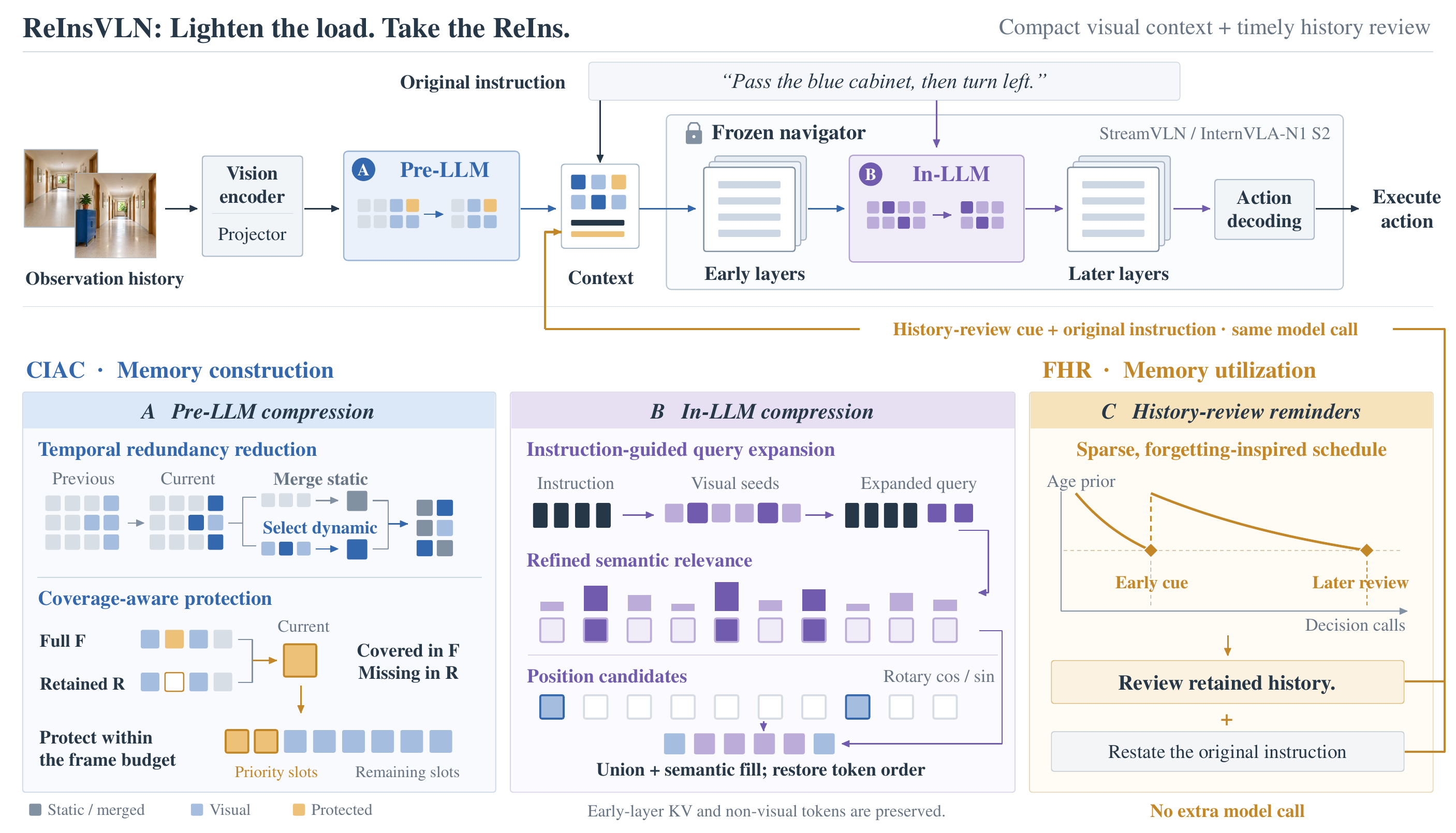}
\caption{Overview of ReInsVLN. CIAC supports memory construction through Pre-LLM redundancy reduction and coverage-aware protection, followed by instruction-guided In-LLM token selection. FHR supports memory utilization by guiding the frozen navigator to revisit retained history through scheduled reminders, without additional model calls.}
\label{fig:reinsvln}
\end{figure*}

\subsection{Behavioral Profiles of History Review}

ReInsBench identifies recoverable continuations after navigation begins (Table~\ref{tab:diagnostic}). On the 148 later checkpoints with all three conditions, $R_0$ recovers 24 of 111 $B$ failures, compared with 17 for $N$. Accounting for ten regressions, $R_0$ gains 9.46 percentage points over $B$ and 2.70 over $N$. Across the 198-instance subset, the corresponding gains are 12.12 and 5.56 points, with recoveries on all three navigators. In the full collection, recoveries relative to $N$ are partly offset by regressions, yielding a net gain of 2.34 points (95\% trajectory-cluster interval $[-0.99,5.68]$).

These outcomes show recovery and disruption, while the full-set confidence interval leaves the advantage over neutral prompting uncertain. Seven of the 198-instance subset's eleven net $R_0-N$ successes arise at initial registration. Figure~\ref{fig:bench_examples} illustrates selected recoveries. Because $R_0$ combines history review with task restatement, attribution to historical evidence alone requires content-matched and history-availability controls. Registration can precede actual prompt consumption, particularly for StreamVLN. The supplement details these boundaries and attribution controls.

\section{ReInsVLN: Visual Compression and Timely Reminders}
\label{sec:reinsvln}

\subsection{Overview}

As shown in Fig.~\ref{fig:reinsvln}, ReInsVLN combines \textbf{Coverage- and Instruction-Aware Compression (CIAC)} and \textbf{Forgetting-Inspired History Reminders (FHR)} for memory construction and utilization. CIAC reduces temporal redundancy and protects content poorly covered by retained tokens in the Pre-LLM stage, then selects visual tokens by instruction relevance and position in the In-LLM stage. FHR inserts scheduled reminders into existing decision calls to prompt history review guided by the original instruction. Both use the frozen navigator's existing context mechanisms, preserving native action decoding and execution.

\subsection{Coverage- and Instruction-Aware Compression (CIAC)}

\noindent \textbf{Pre-LLM Compression.}
% Pre-LLM compression combines causal temporal reduction with dynamic-token protection. Causal temporal reduction compares tokens at corresponding patch positions in the current and preceding uncompressed frames, without accessing future observations. Tokens whose cosine similarity exceeds a threshold are classified as static, and the remainder as dynamic. The fixed frame budget $B_{\mathrm{pre}}$ is allocated proportionally between these groups. Static tokens are clustered and merged into representatives, while dynamic tokens are selected by saliency~\cite{chen2026streamingtom}.
Pre-LLM compression combines causal temporal reduction with dynamic-token protection. Tokens at corresponding patch positions in the current and preceding uncompressed frames are classified as static if their cosine similarity exceeds a threshold, and dynamic otherwise. The frame budget $B_{\mathrm{pre}}$ is allocated proportionally between these groups. Static tokens are clustered and merged, while dynamic tokens are selected by saliency~\cite{chen2026streamingtom}.

The compressed representation of the preceding frame may fail to preserve some original content. Dynamic token protection identifies this coverage gap by comparing current tokens with both the full and retained references. For a current visual token $\mathbf{x}_i$, let $F$ and $R$ denote the full and retained token sets of the preceding frame. Their coverage is measured by maximum cosine similarity:
\begin{equation}
u_i = \max_{\mathbf{f}\in F}\cos(\mathbf{x}_i,\mathbf{f}),
\quad
v_i = \max_{\mathbf{r}\in R}\cos(\mathbf{x}_i,\mathbf{r}).
\label{eq:temporal_coverage}
\end{equation}
Given a similarity threshold $\theta$, tokens satisfying $u_i \geq \theta$ and $v_i < \theta$ resemble full reference content but are insufficiently covered by retained tokens. These candidates are ranked by
\begin{equation}
d_i = \bar{s}_i \bigl[1 + (u_i - v_i)\bigr],
\label{eq:dynamic_priority}
\end{equation}
% where $\bar{s}_i$ denotes normalized token saliency. The score prioritizes salient content with a larger coverage gap. Up to $K$ highest-ranked candidates are retained first, and the remaining budget is allocated to static-token merging and dynamic-token selection, keeping the total within $B_{\mathrm{pre}}$.
where $\theta$ is a similarity threshold and $\bar{s}_i$ is normalized token saliency. Up to $K$ highest-ranked candidates are retained first, with the remaining budget allocated to static-token merging and dynamic-token selection.

\noindent \textbf{In-LLM Compression.}
% The In-LLM stage further selects retained visual tokens according to the original instruction and positional information. At pruning layer $\ell$, let $\mathbf{V}\in\mathbb{R}^{n\times d}$ contain the value representations of one frame's visual tokens, and let $\mathbf{Q}_I\in\mathbb{R}^{m\times d}$ contain those of the original instruction tokens, with value heads concatenated. For a query matrix $\mathbf{Q}$, visual relevance is computed as
The In-LLM stage further selects visual tokens using the original instruction and positional information. At pruning layer $\ell$, let $\mathbf{V}\in\mathbb{R}^{n\times d}$ and $\mathbf{Q}_I\in\mathbb{R}^{m\times d}$ denote the concatenated value-head representations of one frame's visual tokens and the original instruction tokens, respectively. For a query matrix $\mathbf{Q}$, visual relevance is
\begin{equation}
\mathbf{a}(\mathbf{Q}) =
\operatorname{Mean}_{\mathrm{rows}}
\left[
\operatorname{softmax}
\left(
\frac{\mathbf{Q}\mathbf{V}^{\top}}{\sqrt{d}}
\right)
\right],
\label{eq:inllm_relevance}
\end{equation}
% where softmax is applied over visual tokens and scores are averaged over query tokens. The top $k_s$ visual tokens under $\mathbf{a}(\mathbf{Q}_I)$ serve as seeds. Their value representations are appended to $\mathbf{Q}_I$ to form an expanded query $\mathbf{Q}'$, yielding refined scores $\tilde{\mathbf{a}}=\mathbf{a}(\mathbf{Q}')$. This allows instruction-selected visual content to participate in ranking the frame's tokens.
with softmax over visual tokens and averaging over query tokens. The value representations of the top $k_s$ visual tokens under $\mathbf{a}(\mathbf{Q}_I)$ are appended to $\mathbf{Q}_I$, forming an expanded query $\mathbf{Q}'$ with refined scores $\tilde{\mathbf{a}}=\mathbf{a}(\mathbf{Q}')$.

Positional information provides an additional source of candidates. For each visual token, the components of its rotary cosine and sine vectors are summed separately. The indices of the top $k_p$ tokens under each score are combined into a positional candidate set $P$. If this set exceeds the frame budget $B_{\mathrm{in}}$, it is reduced by semantic rank. The remaining slots are filled using $\tilde{\mathbf{a}}$:
\begin{equation}
T = P
\cup
\operatorname{TopK}_{i\notin P}
\left(
\tilde{a}_i,\,
B_{\mathrm{in}}-|P|
\right),
\label{eq:inllm_selection}
\end{equation}
where $\operatorname{TopK}$ returns the indices with
the highest scores, and $T$ denotes the selected
token index set. The selected tokens retain their original positions and sequence order. The shortened hidden sequence is passed to subsequent layers, reducing the tokens they process and cache. Layers up to $\ell$ retain their KV entries, and non-visual tokens are preserved.

\subsection{Forgetting-Inspired History Reminders (FHR)}

Compression determines which visual evidence remains available, but retaining evidence does not ensure its use in later decisions. FHR complements CIAC by reconnecting retained observations with the original instruction. The recoveries and regressions observed in ReInsBench motivate sparse reminders that encourage history utilization while limiting repeated intervention.

% Inspired by forgetting curves and forgetting-based memory refresh~\cite{zhong2024memorybank}, we use an exponential age prior to organize reminder timing. Let $t$ index model decision calls and $t_{\mathrm{last}}$ denote the last reminder call. Before the $k$-th reminder, the age prior is
Inspired by forgetting curves and forgetting-based memory refresh~\cite{zhong2024memorybank}, we schedule reminders using an exponential age prior. Let $t$ index model decision calls and $t_{\mathrm{last}}$ denote the last reminder call. Before reminder $k$, the prior is
\begin{equation}
q(t)=\exp\left[-\frac{t-t_{\mathrm{last}}}{\tau_k}\right],
\label{eq:retention}
\end{equation}

% where $\tau_k > 0$ specifies the reminder interval. A reminder is triggered when $q(t) \leq e^{-1}$, equivalently when $t - t_{\mathrm{last}} \geq \tau_k$. The trigger resets $t_{\mathrm{last}}$ to $t$ and advances the schedule. This formulation expresses a forgetting-inspired temporal heuristic; it does not estimate the navigator's actual recall probability.
where $\tau_k > 0$ specifies the reminder interval. A reminder is triggered when $q(t) \leq e^{-1}$, equivalently when $t - t_{\mathrm{last}} \geq \tau_k$. The trigger resets $t_{\mathrm{last}}$ to $t$ and advances the schedule. This forgetting-inspired temporal heuristic does not estimate the navigator's actual recall probability.

At each trigger, a fixed prompt asks the navigator to review its available history against the original instruction, followed by a restatement of that instruction. The reminder uses only the retained context, without retrieving discarded frames or introducing future observations, reference actions, or route progress labels. It adds text within an existing decision call, preserving the visual compression budget and native action pipeline without requiring an additional model invocation.

\section{Experiments}
\label{sec:simulation}

\subsection{Experimental Setup}
\noindent\textbf{Simulation Benchmark Setup.}
We evaluate ReInsVLN on all 1,839 R2R val-unseen episodes from 11 scenes. Performance is measured by success rate (SR), success weighted by path length (SPL), and navigation error (NE).

\noindent\textbf{Real-World Evaluation Setup.}
We conduct real-world experiments using a Unitree G1 EDU humanoid robot and a Unitree Go2 EDU quadruped robot, each equipped with an external Intel\textregistered RealSense\texttrademark D435i camera on the front. Both platforms connect to an inference server equipped with an NVIDIA\textregistered RTX 4090 GPU through a Wi-Fi access point, streaming visual observations to the server and receiving action commands for execution.

\subsection{Implementation Details}
We evaluate ReInsVLN with two frozen navigators: StreamVLN~\cite{wei2026streamvln} for streaming inference and InternVLA-N1 System~2~\cite{internnav2025,wei2026ground} for non-streaming inference.

% By default, Pre-LLM compression retains $B_{\mathrm{pre}}=128$ of 196 tokens per eligible frame. Temporal similarity and coverage thresholds are both 0.9, with at most $K=8$ dynamic tokens protected within the budget. Static-token merging uses density-peak clustering with a neighborhood size of 7 and a cluster-center weight of 0.6. Dynamic tokens are protected in eligible frames as they stream into StreamVLN's fast branch. In non-streaming InternVLA-N1, the corresponding protection is applied only to current-observation frames.
By default, Pre-LLM compression retains $B_{\mathrm{pre}}=128$ of 196 tokens per eligible frame.  The temporal similarity and coverage thresholds are both 0.9, with at most $K=8$ dynamic tokens protected within this budget.  This protection applies to eligible frames entering StreamVLN's fast branch and only to current-observation frames in InternVLA-N1. Static-token merging uses density-peak clustering with a neighborhood size of 7 and a cluster-center weight of 0.6.

In-LLM pruning retains $B_{\mathrm{in}}=96$ tokens per eligible frame after layer $\ell=20$. For $n$ input tokens, query expansion uses $k_s=\lceil0.01n\rceil$ visual seeds, and positional selection takes $k_p=\lceil0.05n\rceil$ candidates from each cosine and sine ranking. Pruning applies only to Pre-LLM-compressed frames.

% To evaluate whether sparse reminders improve navigation, we adopt a fixed FHR schedule with reminders at the 2nd and 13th model decision calls. No further reminders are issued after the second trigger. Each reminder asks the navigator to revisit its retained observations and restates the original instruction. I t is incorporated into an existing decision call, without requiring additional model invocations.
For FHR, we instantiate the timing heuristic with a fixed schedule, triggering reminders only at the 2nd and 13th model decision calls. Each reminder prompts the navigator to revisit retained observations and restates the original instruction. Reminders are incorporated into existing decision calls without additional model invocations.

\begin{table}[b]
\centering
\caption{Simulation results on R2R Val-Unseen. Changes are relative to the corresponding baseline.}
\label{tab:main}
\setlength{\tabcolsep}{4pt}
\begin{tabular}{l|ccc}
\toprule
\multirow{2}{*}{\textbf{Method}}
& \multicolumn{3}{c}{\textbf{R2R Val-Unseen}} \\
& \textbf{SR (\%) $\uparrow$} & \textbf{SPL (\%) $\uparrow$} & \textbf{NE (m) $\downarrow$} \\
\midrule
\multicolumn{4}{c}{\textbf{Streaming VLN: StreamVLN}} \\
\midrule
Baseline & 55.79 & 49.90 & 4.81 \\
ReInsVLN & \textbf{64.27}\increase{8.48} & \textbf{56.42}\increase{6.52} & \textbf{4.35}\decrease{0.46} \\
\midrule
\multicolumn{4}{c}{\textbf{Non-Streaming VLN: InternVLA-N1}} \\
\midrule
Baseline & 61.07 & 55.67 & 4.26 \\
ReInsVLN & \textbf{64.44}\increase{3.37} & \textbf{58.79}\increase{3.12} & \textbf{4.09}\decrease{0.17} \\
\bottomrule
\end{tabular}
\end{table}

\begin{figure*}[t]
    \centering
    \includegraphics[width=\textwidth]{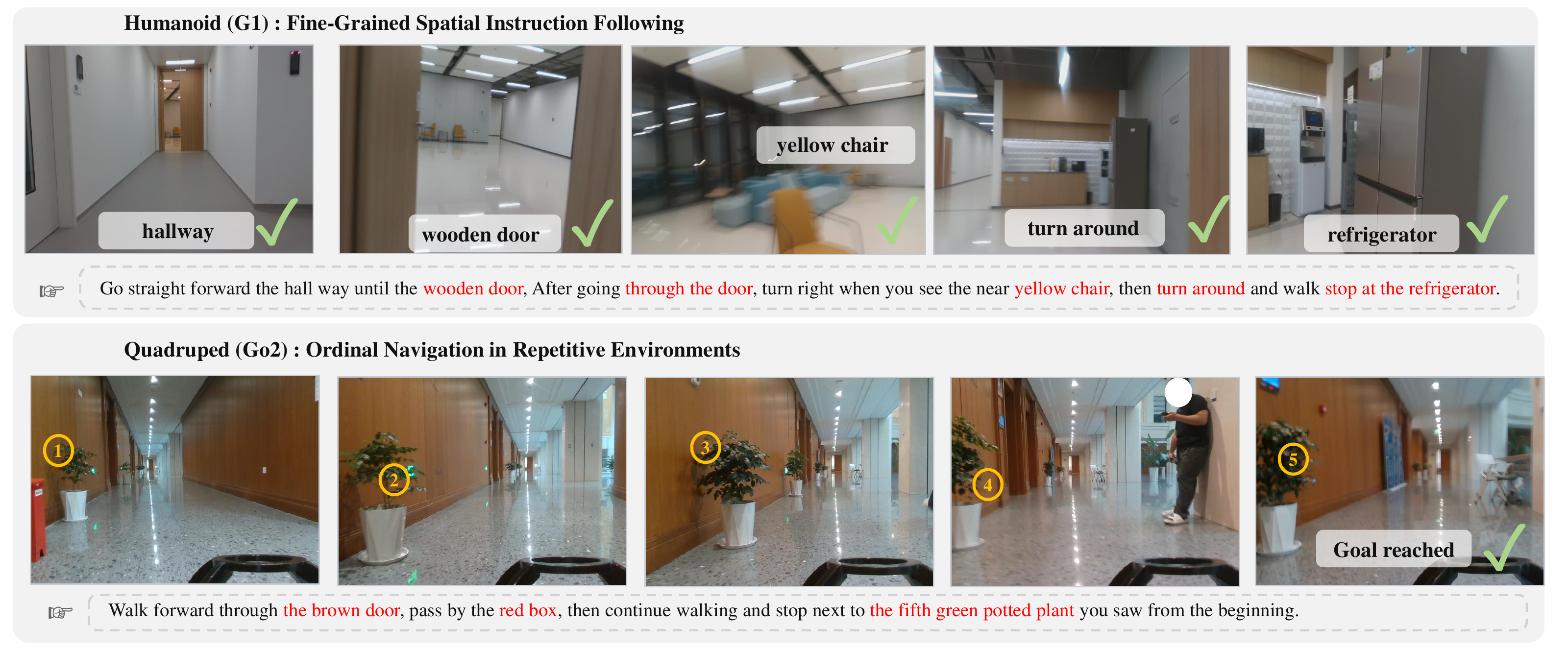}
    \caption{Representative real-world navigation examples.
    Top: the Unitree G1 humanoid follows a sequence of
    landmark-dependent spatial instructions.
    Bottom: the Unitree Go2 quadruped navigates through
    repeated landmarks and stops beside the fifth potted
    plant in encounter order.
    Red text highlights instruction clauses, and numbered
    markers indicate the order of encountered plants.}
    \label{fig:realworld}
\end{figure*}

\subsection{Main Results}

\noindent\textbf{Simulation Experimental Results.}
Table~\ref{tab:main} reports navigation performance on R2R Val-Unseen. ReInsVLN improves SR by 8.48 and 3.37 points over the native StreamVLN and InternVLA-N1, reaching 64.27\% and 64.44\%, respectively. These gains are accompanied by SPL improvements of 6.52 and 3.12 points and NE reductions of 0.46\,m and 0.17\,m. Higher SPL and lower NE complement the success-rate gains, indicating improvements in success-weighted path efficiency and final-goal proximity. The consistent gains across both navigators support the effectiveness of ReInsVLN under streaming and non-streaming inference.

\begin{figure}[!t]
\centering
\includegraphics[width=0.95\linewidth]{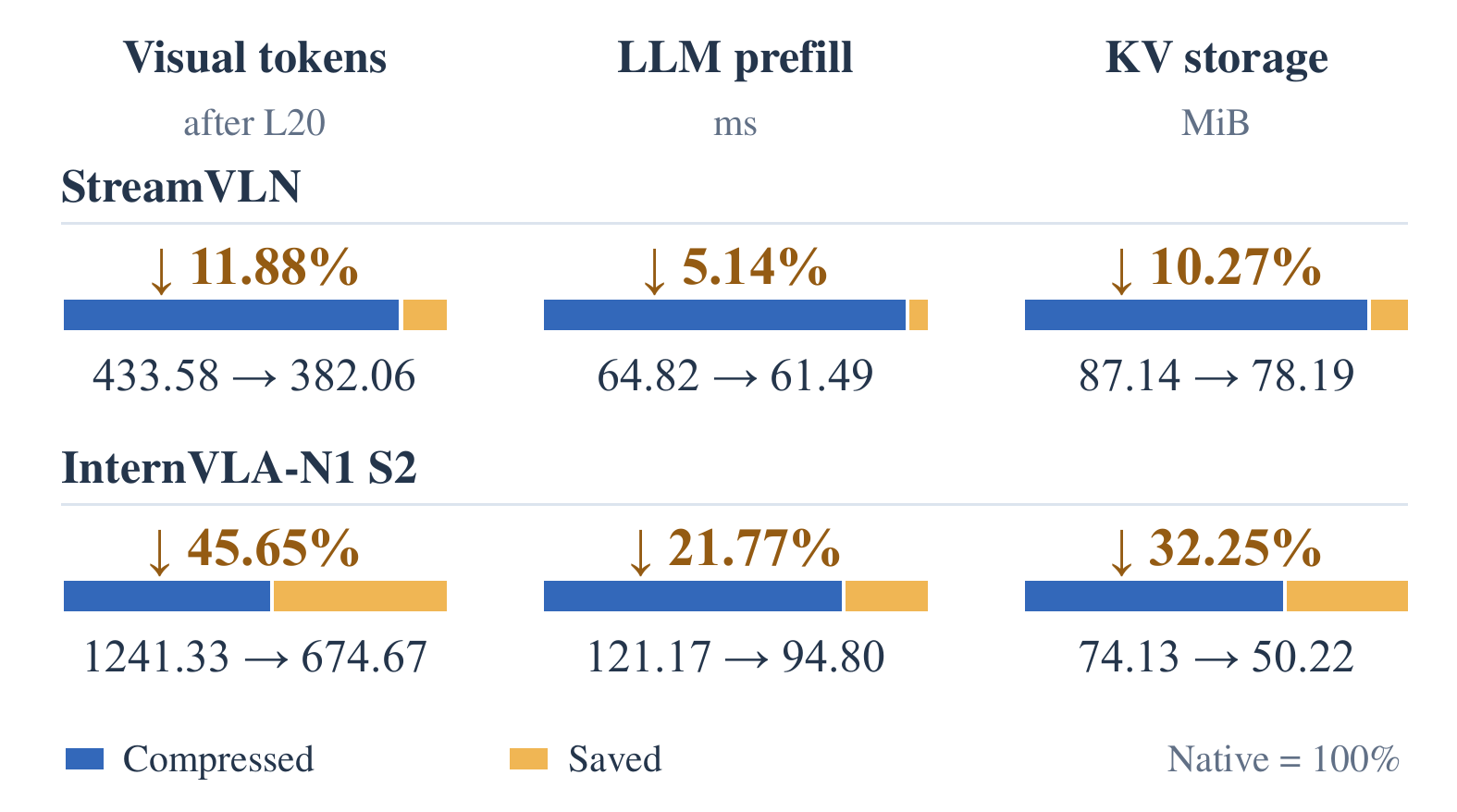}
\caption{Compression efficiency with reminders disabled.
Bars are normalized to Native (100\%): blue indicates
compressed cost and gold indicates savings.
Percentages report relative reductions; values below show
Native $\rightarrow$ compressed means.
Metrics are visual tokens after layer 20, LLM GPU prefill
time, and end-of-call KV tensor storage excluding weights.
Means cover 99 matched normal calls from 33 episodes,
each repeated three times; each condition uses a distinct
dedicated A100 40GB GPU.}
\label{fig:costs}
\end{figure}

\noindent\textbf{{Inference Efficiency.}}
Since pruning ratios alone do not determine practical speedups~\cite{hu2026beyond}, we benchmark CIAC against native inference on 33 episodes across 11 scenes, with FHR disabled. Matched observations are replayed at controlled four-action boundaries. We sample three normal decision calls per episode at early, middle, and late stages, giving 99 matched calls, each repeated three times. InternVLA-N1 look-down continuations are excluded. Each condition uses a dedicated A100 40GB GPU. Figure~\ref{fig:costs} shows that CIAC reduces visual tokens after layer 20, LLM GPU prefill latency, and KV cache memory by 11.88\%, 5.14\%, and 10.27\% on StreamVLN, respectively. InternVLA-N1 exhibits larger corresponding reductions of 45.65\%, 21.77\%, and 32.25\%. This may partly reflect their inference patterns. StreamVLN incrementally processes incoming frames reusing cached history, whereas InternVLA-N1 jointly processes uniformly sampled historical frames and the current observation, providing more visual context for compression per call.

\begin{table}[!t]
\centering
\caption{Real-world results on ordinal navigation with
repeated landmarks using Unitree Go2. The robot is instructed
to stop beside the fifth potted plant. Provisional SR counts
stops between adjacent plants as failures.}
\label{tab:realworld_ordinal}
\small
\setlength{\tabcolsep}{5pt}
\begin{tabular}{lccc}
\hline
Method & Trials & Successes & SR (\%) $\uparrow$ \\
\hline
StreamVLN & 4 w/ & 1 & 25.00 \\
InternVLA-N1 & 4 w/o & 0 & 0.00 \\
\textbf{ReInsVLN} & 4 w/ + 4 w/o & 3 + 2 & \textbf{62.50} \\
\hline
\end{tabular}

\vspace{2pt}
\parbox{\columnwidth}{\footnotesize\raggedright
w/: with doorway crossing; w/o: without doorway crossing. Instructions differ accordingly, but both variants target the fifth potted plant in the same corridor scene.
}
\end{table}

\noindent\textbf{{Real-World Experimental Results.}}
We evaluate ReInsVLN on two real-world tasks: sequential spatial instruction following with the Unitree G1 humanoid and ordinal navigation with the Unitree Go2 quadruped. The latter requires the robot to stop beside the fifth potted plant along a corridor with repeated landmarks, testing its ability to track landmark encounters across observations.

% As shown in Table~\ref{tab:realworld_ordinal}, ReInsVLN succeeds in five of eight ordinal navigation trials (62.5\%), compared with one of four (25.0\%) for StreamVLN and zero of four for InternVLA-N1.  These results suggest that better utilization of retained history can improve the real-world generalization of frozen navigators, helping them handle tasks requiring sequential landmark tracking that remain challenging for their native counterparts. Figure~\ref{fig:realworld} presents representative navigation sequences for both tasks.
As shown in Table~\ref{tab:realworld_ordinal}, ReInsVLN succeeds in five of eight ordinal navigation trials (62.5\%), compared with one of four (25.0\%) for StreamVLN and zero of four for InternVLA-N1. These results suggest that better utilization of retained history can improve frozen navigators' real-world generalization, helping them handle sequential landmark tracking tasks that challenge their native counterparts. Figure~\ref{fig:realworld} presents representative navigation sequences for both tasks.

% \begin{figure*}[!t]
%     \centering
%     \includegraphics[width=\textwidth]{figures/real_experi.pdf}
%     \caption{Representative real-world navigation examples.
%     Top: the Unitree G1 humanoid follows a sequence of
%     landmark-dependent spatial instructions.
%     Bottom: the Unitree Go2 quadruped navigates through
%     repeated landmarks and stops beside the fifth potted
%     plant in encounter order.
%     Red text highlights instruction clauses, and numbered
%     markers indicate the order of encountered plants.}
%     \label{fig:realworld}
% \end{figure*}

\begin{table}[t]
\caption{Component ablation on R2R Val-Unseen. Shaded rows indicate the settings used in ReInsVLN.}
\label{tab:component}

{\centering
\begin{tabular}{c|cc|ccc}
\toprule
\multirow{2}{*}{\#}
& \multicolumn{2}{c|}{\textbf{Components}}
& \multicolumn{3}{c}{\textbf{R2R Val-Unseen}} \\
& \textbf{CIAC} & \textbf{FHR}
& \textbf{SR (\%) $\uparrow$}
& \textbf{SPL (\%) $\uparrow$}
& \textbf{NE (m) $\downarrow$} \\
\midrule
\multicolumn{6}{c}{\textbf{Streaming VLN: StreamVLN}} \\
\midrule
1 & & & 55.79 & 49.90 & 4.81 \\
2 & \checkmark & &
56.77\increase{0.98} &
50.71\increase{0.81} &
4.83\increase{0.02} \\
3 & & \checkmark &
60.96\increase{5.17} &
54.33\increase{4.43} &
4.62\decrease{0.19} \\
\rowcolor{gray!20}
4 & \checkmark & \checkmark &
\textbf{64.27}\increase{8.48} &
\textbf{56.42}\increase{6.52} &
\textbf{4.35}\decrease{0.46} \\
\midrule
\multicolumn{6}{c}{\textbf{Non-Streaming VLN: InternVLA-N1}} \\
\midrule
5 & & & 61.07 & 55.67 & 4.26 \\
6 & \checkmark & &
60.74\decrease{0.33} &
55.78\increase{0.11} &
4.29\increase{0.03} \\
7 & & \checkmark &
64.06\increase{2.99} &
58.05\increase{2.38} &
\textbf{4.09}\decrease{0.17} \\
\rowcolor{gray!20}
8 & \checkmark & \checkmark &
\textbf{64.44}\increase{3.37} &
\textbf{58.79}\increase{3.12} &
\textbf{4.09}\decrease{0.17} \\
\bottomrule
\end{tabular}
\par}

% \vspace{2pt}
% \noindent\makebox[\linewidth][c]{%
%   \parbox{0.8\linewidth}{%
%     \footnotesize\raggedright
%     CIAC uses the shaded settings in Table~VI.
%     \par
%   }%
% }
% \par

\end{table}

\subsection{Ablation Studies}

\noindent\textbf{Component Ablation.}
ReInsVLN consists of two components: Coverage- and Instruction-Aware Compression (CIAC) and Forgetting-Inspired History Reminders (FHR). To explore the contribution of each component to our navigator, we conduct ablation studies on different component combinations. The results are summarized in Table~\ref{tab:component}. CIAC substantially reduces visual tokens while largely preserving navigation performance. Its two-stage compression reduces the token budget per eligible frame from 196 to 128 and then to 96. Despite this reduction, SR remains close to the baseline, with changes of +0.98\% and -0.33\% on StreamVLN and InternVLA-N1, respectively, while SPL increases by 0.81\% and 0.11\%. FHR alone improves SR by 5.17 and 2.99 percentage points on StreamVLN and InternVLA-N1, respectively, with higher SPL and lower NE. It recovers 173 and 126 baseline failures, while introducing 78 and 71 success-to-failure regressions. The recoveries outweigh the regressions on both navigators, producing net gains of 95 and 55 successful episodes, respectively. Combining CIAC and FHR achieves the highest reported SR and SPL among the evaluated configurations on both navigators. Compared with FHR alone, SR increases by a further 3.31\% and 0.38\% on StreamVLN and InternVLA-N1, respectively.

\begin{table}[!b]
\centering
\caption{Ablation of protection strategies and targets on R2R Val-Unseen. Stat. and Dyn. denote static and dynamic tokens, respectively. Curr. and Hist. refer to the current and historical frames, respectively. In non-streaming InternVLA-N1, current-frame protection corresponds to fast-branch protection in StreamVLN.}
\label{tab:proj_comp}
\begin{tabular}{c|cc|ccc}
\toprule
\multirow{2}{*}{\#}
& \multicolumn{2}{c|}{\textbf{Protection}}
& \multicolumn{3}{c}{\textbf{R2R Val-Unseen}} \\
& \textbf{Method} & \textbf{Target} & \textbf{SR (\%) $\uparrow$} & \textbf{SPL (\%) $\uparrow$} & \textbf{NE (m) $\downarrow$} \\
\midrule
\multicolumn{6}{c}{\textbf{Streaming VLN: StreamVLN}} \\
\midrule
1 & w/o prot. & - & 56.77 & 50.48 & 4.79 \\
2 & Saliency & Dyn. & 56.99 & 50.79 & 4.76 \\
\midrule
3 & Coverage & Stat.  & 55.79 & 49.71 & 4.83 \\
\rowcolor{gray!20}
4 & Coverage & Dyn. & \textbf{57.31} & \textbf{50.81} & \textbf{4.73} \\
5 & Coverage & Stat. \& Dyn. & 55.14 & 49.48 & 4.89 \\
\midrule
\multicolumn{6}{c}{\textbf{Non-Streaming VLN: InternVLA-N1}} \\
\midrule
6 & w/o prot. & -    & \textbf{61.45} & \textbf{55.94} & 4.32 \\
7 & Saliency & Dyn.  & 60.74 & 55.48 & 4.35 \\
\midrule
8 & Coverage & Stat.            & 60.58 & 55.47 & 4.30 \\
\rowcolor{gray!20}
9 & Coverage & Dyn.           & 61.01 & 55.71 & 4.34 \\
10 & Coverage & Stat. \& Dyn. & 59.92 & 54.91 & 4.30 \\
\midrule
\rowcolor{gray!20}
11 & Coverage & Curr. Dyn. & 61.23 & 55.65 & \textbf{4.24} \\
12 & Coverage & Hist. Dyn. & 60.79 & 55.60 & 4.31 \\
\bottomrule
\end{tabular}
% \par\vspace{2pt}
% {\footnotesize\raggedright
% Pre-LLM: 128 tokens/frame; In-LLM pruning and FHR off.
% \par}
\end{table}

\noindent\textbf{Coverage-Aware Protection.}
Table~\ref{tab:proj_comp} compares protection strategies and targets within CIAC. On StreamVLN, coverage-based protection of dynamic tokens outperforms saliency-based protection and other target configurations. Compared with no protection, it improves SR and SPL by 0.54\% and 0.33\%, respectively, and reduces NE by 0.06\,m. The limited benefit on InternVLA-N1 may reflect larger temporal gaps from uniform frame sampling, potentially weakening the inter-frame correspondence used by coverage-based protection. Current-frame dynamic-token protection, corresponding to fast-branch protection in StreamVLN, achieves higher SR and lower NE than protection over all frames or historical frames. Compared with no protection, NE decreases from 4.32 to 4.24\,m, while SR and SPL decrease slightly by 0.22\% and 0.29\%, respectively. These results suggest adapting the protection scope to each navigator.

\noindent\textbf{Keep Ratio, Pruning Layer, Scope, and Early KV Pruning.}
Table~\ref{tab:llm_comp} evaluates in-LLM compression settings, prioritizing navigation performance preservation. A 0.75 keep ratio retains 96 of 128 tokens per eligible frame. Reducing it to 0.50 degrades performance on InternVLA-N1 but yields mixed results on StreamVLN. With other settings at their defaults, moving pruning from layer 16 to 20 improves SR by 1.90\% and 1.31\% on StreamVLN and InternVLA-N1, respectively, alongside higher SPL and lower NE. \textit{Proj.} targets frames already compressed at the projector stage, while \textit{All} targets all frames. Under otherwise default settings, \textit{Proj.} improves all metrics on StreamVLN and SPL on InternVLA-N1, with slightly worse SR and NE on the latter.

We further evaluate early KV pruning, which additionally removes the KV cache entries of discarded visual tokens from all layers up to the pruning layer. Adding it to the default configuration reduces SR by 0.71\% and 0.27\% on StreamVLN and InternVLA-N1, respectively, while also lowering SPL and increasing NE.

\begin{table}[t]
\centering
\caption{Ablation of in-LLM compression settings on R2R Val-Unseen.}
\label{tab:llm_comp}
\resizebox{\columnwidth}{!}{
\begin{tabular}{c|cccc|ccc}
\toprule
\multirow{2}{*}{\#}
& \multirow{2}{*}{\textbf{\makecell{Pruning\\Layer}}}
& \multirow{2}{*}{\textbf{\makecell{Keep\\Ratio}}}
& \multirow{2}{*}{\textbf{Scope}}
& \multirow{2}{*}{\textbf{\makecell{Early KV\\Pruning}}}
& \multicolumn{3}{c}{\textbf{R2R Val-Unseen}} \\
& & & & & \textbf{SR (\%) $\uparrow$} & \textbf{SPL (\%) $\uparrow$} & \textbf{NE (m) $\downarrow$} \\
\midrule
\multicolumn{8}{c}{\textbf{Streaming VLN: StreamVLN}} \\
\midrule
1 & 16 & .75 & Proj. & $\times$   & 54.87 & 48.57 & 5.02 \\
2 & 16 & .75 & Proj. & \checkmark & 56.33 & 50.09 & \textbf{4.80} \\
3 & 16 & .50 & Proj. & $\times$   & 55.30 & 49.11 & 5.02 \\
4 & 16 & .50 & Proj. & \checkmark & 55.95 & 49.51 & 4.88 \\
\midrule
\rowcolor{gray!20}
5 & 20 & .75 & Proj. & $\times$   & \textbf{56.77} & \textbf{50.71} & 4.83 \\
6 & 20 & .75 & Proj. & \checkmark & 56.06 & 49.77 & 4.95 \\
7 & 20 & .75 & All   & $\times$   & 56.17 & 50.22 & 4.94 \\
8 & 20 & .75 & All   & \checkmark & 56.06 & 50.03 & 4.98 \\
\midrule
\multicolumn{8}{c}{\textbf{Non-Streaming VLN: InternVLA-N1}} \\
\midrule
9 & 16 & .75 & Proj.  & $\times$   & 59.43 & 54.66 & 4.44 \\
10 & 16 & .75 & Proj. & \checkmark & 60.30 & 55.48 & 4.37 \\
11 & 16 & .50 & Proj. & $\times$   & 58.18 & 53.43 & 4.48 \\
12 & 16 & .50 & Proj. & \checkmark & 58.51 & 53.76 & 4.45 \\
\midrule
\rowcolor{gray!20}
13 & 20 & .75 & Proj. & $\times$   & 60.74 & \textbf{55.78} & 4.29 \\
14 & 20 & .75 & Proj. & \checkmark & 60.47 & 55.30 & 4.31 \\
15 & 20 & .75 & All   & $\times$   & \textbf{60.90} & 55.41 & \textbf{4.27} \\
16 & 20 & .75 & All   & \checkmark & 59.82 & 54.73 & 4.28 \\
\bottomrule
\end{tabular}
}
\end{table}

\newpage
\section{Conclusion}

We presented ReInsVLN, a training-free framework for efficient embodied navigation through compact visual context and improved history utilization. ReInsBench examines history utilization through controlled prompting comparisons, motivating the combination of coverage- and instruction-aware compression with scheduled history-review reminders. Experiments on R2R Val-Unseen show improved navigation performance on StreamVLN and InternVLA-N1, while visual compression reduces prefill latency and KV cache storage. These findings support visual context compression and history review as complementary approaches to improving navigation performance and reducing inference overhead in embodied systems.

\bibliographystyle{IEEEtran}
\bibliography{references}
\end{document}